\documentclass{article}
\usepackage{spconf,amsmath,graphicx}

\usepackage{graphicx}
\usepackage{comment}
\usepackage{amsmath,amssymb} % define
\usepackage{color}
\usepackage{amsmath}
\usepackage{amssymb}
\usepackage{subfigure}
\usepackage{booktabs} 
\usepackage{graphicx}
\usepackage{paralist}
\usepackage{xcolor}
\usepackage{booktabs}  
\usepackage{multirow} 
\usepackage{multicol} 
\usepackage{arydshln}
\usepackage{verbatim}
\usepackage{arydshln}
\usepackage{mathtools}
\usepackage{verbatim}

\newsavebox\CBox
\def\textBF#1{\sbox\CBox{#1}\resizebox{\wd\CBox}{\ht\CBox}{\textbf{#1}}}

\newlength\secmargin
\newlength\paramargin
\newlength\figmargin
\title{Learning an evolved mixture model for task-free continual learning}
\name{Fei Ye and Adrian G. Bors}
\address{Department of Computer Science, University of York, York YO10 5GH, UK}
\begin{document}
%\ninept
%
\maketitle
\begin{abstract}
Recently, continual learning (CL) has gained significant interest because it enables deep learning models to acquire new knowledge without forgetting previously learnt information. However, most existing works require knowing the task identities and boundaries, which is not realistic in a real context. In this paper, we address a more challenging and realistic setting in CL, namely the Task-Free Continual Learning (TFCL) in which a model is trained on non-stationary data streams with no explicit task information. To address TFCL, we introduce an evolved mixture model whose network architecture is dynamically expanded to adapt to the data distribution shift. We implement this expansion mechanism by evaluating the probability distance between the knowledge stored in each mixture model component and the current memory buffer using the Hilbert Schmidt Independence Criterion (HSIC). We further introduce two simple dropout mechanisms to selectively remove stored examples in order to avoid memory overload while preserving memory diversity. Empirical results demonstrate that the proposed approach achieves excellent performance. 
\end{abstract}
\section{Introduction}
\label{sec:intro}
Continual learning, also called lifelong learning, is one of the essential functions in an artificial intelligence system, representing the ability to continually remember the entire previously learnt experiences from a sequence of tasks \cite{LifeLong_review}. Such abilities are inherited in humans and animals, enabling them to survive in the dynamically changing environment during their entire life. However, deep learning systems would usually perform well on individual tasks \cite{InfoVAEGAN_conference,JontLatentVAEs} but suffer from dramatic performance loss when training on several different tasks sequentially \cite{LifelongVAEGAN,Lifelonginterpretable,LifelongTwin}. The reason behind the performance loss is the network forgetting when its parameters are replaced following the training with a new task \cite{LifeLong_review}.  

In this paper, we address a more challenging and realistic learning setting in CL, called Task-Free Continual Learning (TFCL), which assumes that the task identities and boundaries are not available during the training. One popular attempt for TFCL is to employ a small memory buffer to store incoming samples at each training step \cite{ContinualPrototype}. Such an approach performs well on TFCL tasks when its memory buffer contains diverse data samples \cite{ContinualPrototype}. However, there are two main drawbacks for the memory-based approaches: \begin{inparaenum}[1)]
\item The model would suffer from the negative backward transfer when the memory buffer stores incoming samples which are sufficiently different those learnt previously;
\item It can not address infinite data streams due to the fixed memory capacity.
\end{inparaenum} In this paper, we address these drawbacks by introducing the Evolved Mixture Model (EEM) which is a continual learning model avoiding the negative backward transfer by adding additional model capacity for learning incoming samples. First, we implement each expert in EEM by using a VAE model for the model selection and a classifier for the prediction task. We then introduce two simple dropout mechanisms for regularizing the memory buffer by selectively removing stored samples to avoid memory overload. We further introduce a new expansion mechanism that evaluates the Hilbert Schmidt Independence Criterion (HSIC) between the information stored in each expert and the current memory as a mixture expansion signal. The proposed expansion mechanism enables training a diversity of experts, improving the generalization performance. The other advantage of the proposed HSIC-based expansion mechanism is allowing to perform the unsupervised learning task without requiring any class labels.

The following contributions are brought in this paper:
\vspace*{-0.2cm}
\begin{itemize}
\setlength{\itemsep}{0pt}
\setlength{\parsep}{2pt}
\setlength{\parskip}{2pt}
	\item 
    A new continual learning framework, namely the Evolved Mixture Model (EEM) that can learn an infinite number of data streams without forgetting. 
\item 
    Two simple dropout mechanisms that selectively remove stored samples from memory in order to avoid memory overload. 
\item  
    A new expansion mechanism that utilizes the HSIC criterion to detect the data distribution shift, providing better expansion signals for EEM. 
\end{itemize}

\section{Related work}
\label{RelWork}

\textBF{General continual learning}~: Most existing works focus on the general continual learning, which describes a learning paradigm where a series of tasks is presented, and the model requires recognizing all samples after the training. Existing works for general continual learning can be divided into three branches: regularization-based \cite{LessForgetting}, generative replay mechanism \cite{LifelongVAEGAN,LifelongTeacherStudent} and dynamic expansion approaches \cite{LifelongInfinite,LifelongMixuteOfVAEs}. The regularization-based approaches aim to minimize any change on network's weights that are important to past tasks when training on a novel task \cite{EWC}. The regularization-based approaches still require managing a small memory buffer to store a few past samples used to penalize changes in important network's parameters. Such approaches, however, require a significant computation processing when learning a long sequence of tasks, \cite{AGEM}. The Generative Replay Mechanism (GRM) usually requires training a generator model, such as a Variational Autoencoder (VAE) \cite{VAE} or a Generative Adversarial Network (GAN) as the generative replay network. The dynamic models usually would add new layers with hidden nodes \cite{Adanet} within a single neural network or a task-specific module into a mixture system \cite{LifelongInfinite,NeuralDirichelt,LifelongUnsupervisedVAE}. The former approach is suitable for learning a series of tasks for a single domain while the latter is good when aiming to solve an infinite number of tasks \cite{LifelongInfinite}.

\noindent  \textBF{Task-free continual learning (TFCL)}~: The task-free continual learning (TFCL) can be seen as a special setting in continual learning, which assumes that there are no task boundaries during the training \cite{ContinualPrototype,taskFree_CL}. One of the widespread attempts for TFCL is based on a small memory buffer that aims to store a few past samples to relieve forgetting \cite{ContinualPrototype}. Such an approach requires designing an effective sample selection procedure that selectively stores past samples \cite{ContinualPrototype}. Another approach for TFCL is based on the expansion of the network architecture or by adding new components to a mixture model \cite{DeepMixtureVAE,MixtureOfVAEs,LifelonGGraph}. The approach from \cite{LifelongUnsupervisedVAE} dynamically builds new inference models into a VAE mixture framework when detecting a data distribution shift. GRM was used to relieve forgetting in an approach called Continual Unsupervised Representation Learning (CURL) \cite{CURL}. However, CURL still suffers from forgetting due to the frequent updating of the generator. This issue can be solved by only employing a mixture expansion mechanism only such as the Continual Neural Dirichlet Process Mixture (CN-DPM) \cite{NeuralDirichelt}, which employs Dirichlet processes for expanding the number of VAE components. Although these expansion-based approaches show promising results in TFCL, they do not consider the previously learnt knowledge when performing the expansion. 

\section{Methodology}
\label{sec:proposedMethod}

In this paper, we introduce an evolving mixture model addressing TFCL and in the following section we introduce our approach in detail. The mixture deep learning model and its network architecture are described in Section~\ref{sec:ensemble}. Then in Section~\ref{sec:expansion} we introduce a new model mixture expansion criterion based on the Hilbert Schmidt Independence Criterion (HSIC). Finally, in Section~\ref{sec:dropout} we introduce a new dropout mechanism that removes certain selected data samples to avoid memory overload. 

\subsection{Evolved Mixture Model (EMM)}
\label{sec:ensemble}

Before introducing the proposed approach, we firstly provide the definition of TFCL as follows. Let us consider a set of $n$ training steps, $\mathcal{T} = \{ {\mathcal{T}}_1, {\mathcal{T}}_2, \cdots, {\mathcal{T}}_n \}$ for learning a data stream $\mathcal{D}$. Let ${\mathcal{D}}_i = \{ {\bf x}_{i,j},y_{i,j} \}^b_{j=1}$ represent a small batch of samples drawn from $\mathcal{D}$ at the training step (${\mathcal{T}}_i$), where $b$ is the batch size. Then the data stream $\mathcal{D}$ can be represented by combining all data batches~:
\begin{equation}
\begin{aligned} 
{\mathcal{D}} = \bigcup\limits_{i = 1}^n  {\mathcal{D}}_i \,.
\end{aligned}
\end{equation}
At the $j$-th training step (${\mathcal{T}}_j$), the model only accesses $\mathcal{D}_j$ while all previously visited batches are not available. 

We consider that each component of the mixture consists of a VAE and a classifier. The main motivation for considering the VAE model as expert is that we can perform the selection process based on the sample log-likelihood estimated by the VAE when choosing an appropriate expert during the testing phase. VAEs also provide the latent code which is used for the evaluation of the model's expansion. Let $f_{\xi_i} \colon \mathcal{X} \to \mathcal{Y}$ be a classifier with the trainable parameters ${\xi_i}$ where ${\mathcal{Y}}$ is the space of the model's prediction. In order to avoid frequently building new experts during the training, we introduce a memory buffer, denoted as ${\mathcal{G}}_j$ updated at the training step (${\mathcal{T}}_j$). The loss for the VAE model representing the $K$-th expert is to maximize the marginal log-likelihood \cite{VAE}, defined as~:
\begin{equation}
\begin{aligned}
{\mathcal{L}}^{K}_{VAE}({\mathcal{G} }_j)  \buildrel \Delta \over =  \; & {\mathbb{E}_{{q_{{\omega _K}}}({\bf{z}} \,|\, {\bf{x}})}}\left[ {\log {p_{{\theta _K}}}({\bf{x}} \,|\, {\bf{z}})} \right] \\&- {D_{KL}}\left[ {{q_{{\omega _K}}}({\bf{z}} \,|\, {\bf{x}}) \,||\, p({\bf{z}})} \right],
\label{VAEloss_eq}
\end{aligned}
\end{equation}
where ${p_{{\theta _K}}}({\bf{x}} \,|\, {\bf{z}})$ and ${q_{{\omega _K}}}({\bf{z}} \,|\, {\bf{x}})$ represent the decoding and encoding distributions, which are implemented by two neural networks, respectively. The subscript $K$ represents the index of the expert in the mixture model. We also define the training loss for the classifier on $\mathcal{G}_j$ at the training step (${\mathcal{T}}_j$)~:
\begin{equation}
\begin{aligned}
{\mathcal{L}}^K_{class}({\mathcal{G}}_j)  \buildrel \Delta \over =  \frac{1}{{| {\mathcal{G}}_j |}}\sum\limits_{t = 1}^{|{\mathcal{G}}_j|} \big\{ {\mathcal{L}}_{CE} \big( f_{\xi_K }( {\bf x}_t), { y}_t \big) \big\}  \,,
\label{classifierLoss_eq}
\end{aligned}
\end{equation}
where ${\mathcal{L}}_{CE}(\cdot)$ is the cross-entropy loss function, which is applied on the all data $|{\mathcal{G}}_j|$ from the memory buffer. 

\subsection{Model expansion}
\label{sec:expansion}

We firstly introduce the Hilbert Schmidt Independence Criterion (HSIC) and then describe how this can be used as an expansion mechanism of the mixture model. Let ${\mathcal{Z}}_1$ and $\mathcal{Z}_2$ represent two domains and ${\mathbb{P}}_{{\bf z}_1, {\bf z}_2 }$ be a joint distribution from which we draw a pair of samples $\{{\bf z}_1, {\bf z}_2 \}$ over ${\mathcal{Z}}_1 \times {\mathcal{Z}}_2$. The main goal of HSIC \cite{HSIC} is to measure the independence between ${\bf z}_1$ and ${\bf z}_2$ by evaluating the norm of the cross-covariance operator over the domain ${\mathcal{Z}}_1 \times {\mathcal{Z}}_2$ in reproducing kernel Hilbert space (RKHS) \cite{KernelRKHS}. Let $Q$ and $S$ be the RKHSs on ${\mathcal{Z}}_1$ and ${\mathcal{Z}}_2$ and $f_Q \colon {\mathcal{Z}}_1 \to Q$, $f_S \colon {\mathcal{Z}}_2 \to S$ be their feature functions. We define the associated reproducing kernels as $k({\bf z}_1, {\bf z}'_1) = \left\langle { f_Q({\bf z}_1), f_Q({\bf z}'_1) } \right\rangle$ and $l({\bf z}_2, {\bf z}'_2) = \left\langle { f_S({\bf z}_2), f_S({\bf z}'_2) } \right\rangle$ where ${\bf z}_1, {\bf z}'_1 \in {\mathcal{Z}}_1$ and ${\bf z}_2, {\bf z}'_2 \in {\mathcal{Z}}_2$. The cross-covariance operator between $f_Q$ and $f_S$ is defined as~:
\begin{equation}
\begin{aligned}
C_{{{\bf{z}}_1}{{\bf{z}}_2}} = & \; {{\mathbb{E}}_{{{\bf{z}}_1}{{\bf{z}}_2}}}\Big\{ \left( {{f_Q}({{\bf{z}}_1}) - {{\mathbb{E}}_{{{\bf{z}}_1}}}\left[ {{f_Q}({{\bf{z}}_1})} \right]} \right) \otimes 
\\&
\left( {{f_S}({{\bf{z}}_2}) - {{\mathbb{E}}_{{{\bf{z}}_2}}}\left[ {{f_S}({{\bf{z}}_2})} \right]} \right) \Big\}\,,
\end{aligned}
\end{equation}
where $\otimes$ is the tensor product. Then HSIC is defined as the square of the Hilbert-Schmidt norm of $C_{{\bf z}_1,{\bf z}_2}$~:
\begin{align}
\label{hsic_measure}
{\mathcal{L}}_{HSIC}(Q,S, {\mathbb{P}}_{{\bf z}_1, {\bf z}_2}) &= \left\| { C_{{\bf z}_1, {\bf z}_2} } \right\|_{HS}^2 \notag \\&=
{\mathbb{E}}_{{\bf z}_1,{\bf z}'_1,{\bf z}_2, {\bf z}'_2} [k({\bf z}_1,{\bf z}'_1) l({\bf z}_2,{\bf z}'_2) ] \\&+ {\mathbb{E}}_{{\bf z}_1, {\bf z}'_1}[k({\bf z}_1, {\bf z}'_1 ) ] {\mathbb{E}}_{{\bf z}_2, {\bf z}'_2} [l({\bf z}_2, {\bf z}'_2) ] \notag \\&-
2{\mathbb{E}}_{{\bf z}_1,{\bf z}_2}[{\mathbb{E}}_{{\bf z}'_1}[k({\bf z}_1,{\bf z}'_1 ) ] {\mathbb{E}}[l({\bf z}_2, {\bf z}'_2) ]   ]\,, \notag
\end{align}
\noindent where ${\mathbb{E}}_{{\bf z}_1, {\bf z}'_1, {\bf z}_2, {\bf z}'_2 }$ represents the expectation over paired samples $({\bf z}_1, {\bf z}_2 )$ and $({\bf z}'_1, {\bf z}'_2 )$ drawn from ${\mathbb{P}}_{{\bf z}_1, {\bf z}_2}$.

\begin{figure}[tt]
    \centering
	\includegraphics[scale=0.37]{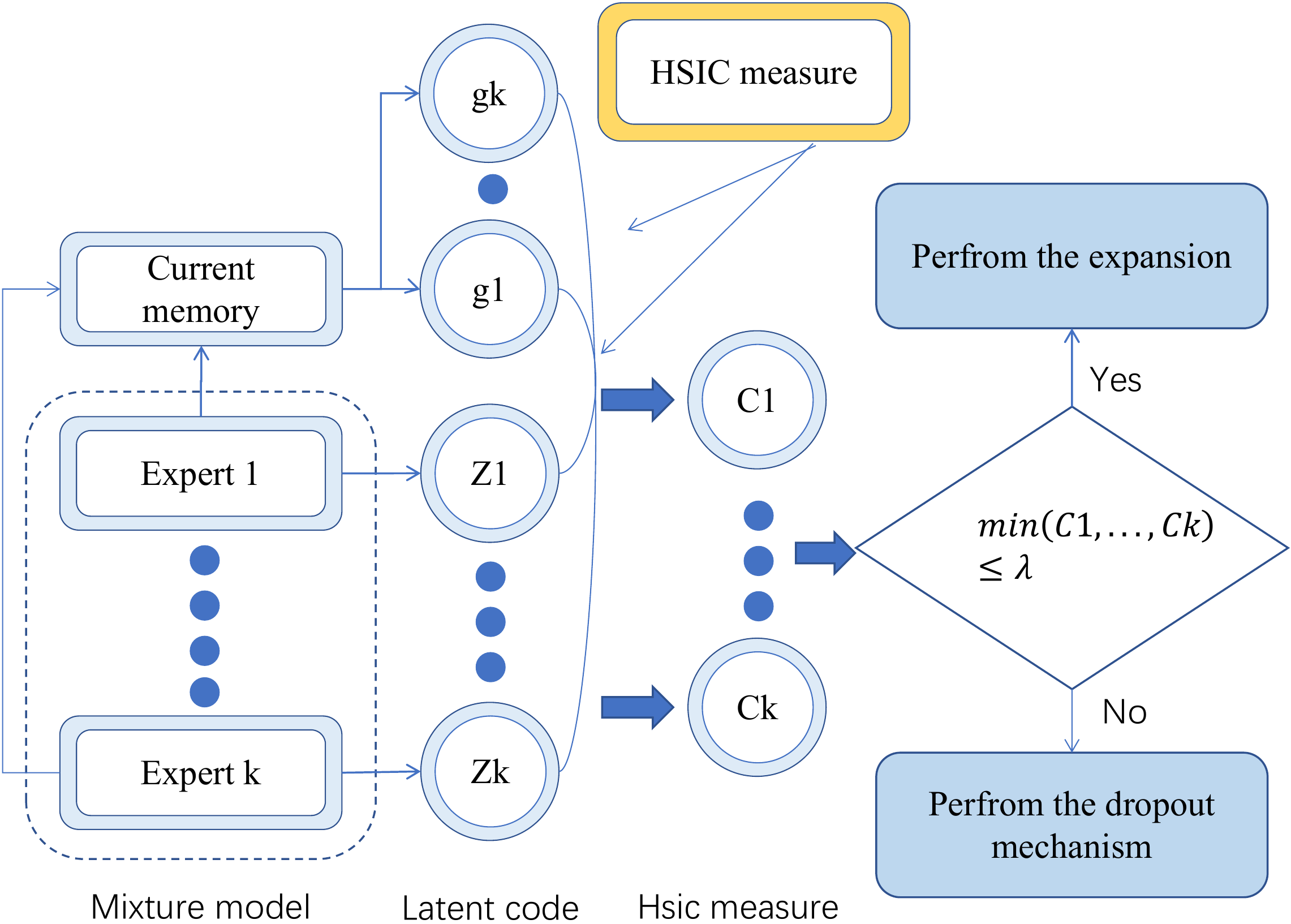}
	\caption{Expansion and dropout mechanism in the proposed EMM model. We first infer the latent variables using each expert and then use HSIC to evaluate the discrepancy between each expert and the current memory. Then we perform either the mixture expansion or the memory buffer data dropout mechanism, according to the expansion criterion, Eq.~\eqref{expansion_criterion}.
  }
	\label{HSICnetwork}
	\vspace{-10pt}
\end{figure}

In the following, we show how HSIC can be used as the expansion criterion for the proposed mixture model. As exemplified in Fig.~\ref{HSICnetwork}, we assume that at  $\mathcal{T}_j$ we have trained $K$ experts in the mixture model. The main idea of the proposed expansion criterion is that if the current data from the memory buffer is novel to the knowledge already accumulated in the trained components, we should build a new component which would learn the new information. Such a mechanism can encourage each component to learn a different underlying data distribution. Let ${\mathbb{P}}_{{\tilde{\bf x}}_i}$ represent the distribution of generative replay samples drawn from the VAE model of the $i$-th expert. Let ${\mathbb{P}}_{{\tilde{\bf z}}_i}$ represent the distribution of the latent variables ${\bf z}$ inferred using the inference model of the $i$-th expert with samples ${\bf x}$ drawn from ${\mathbb{P}}_{{\tilde{\bf x}}_i}$. Let $\mathbb{P}_{g_{j,i}}$ represent the distribution of the latent variables inferred by the $i$-th expert from the stored samples in the memory ${\mathcal{G}}_j$. Let ${\mathbb{P}}_{{\tilde{\bf z}}_i,{g_{j,i}}}$ represent the joint distribution with the marginals  $\mathbb{P}_{g_{j,i}}$ and ${\mathbb{P}}_{{\tilde{\bf z}}_i}$, respectively. Then we estimate HSIC between the knowledge learnt by the $i$-th expert and the distribution of the memory buffer at $\mathcal{T}_j$ by $\mathcal{L}_{HSIC}(Q,S,{\mathbb{P}}_{{\tilde{\bf z}}_i,{g_{j,i}} })$. The expansion criterion for the mixture model at $\mathcal{T}_j$ is defined as~:
\begin{equation}
\begin{aligned}
\lambda <  \min & \{ \mathcal{L}_{HSIC}(Q,S,{\mathbb{P}}_{{\tilde{\bf z}}_1,{g_{j,i}} }), \cdots, \\& \mathcal{L}_{HSIC}(Q,S,{\mathbb{P}}_{{\tilde{\bf z}}_{K-1},{g_{j,K-1}} })   \}\,,
\label{expansion_criterion}
\end{aligned}
\end{equation}
where $\lambda$ is a pre-defined threshold that controls the model's expansion. If Eq.~\eqref{expansion_criterion} holds, we add a new expert to the mixture model. 
The evaluation of Eq.~\eqref{expansion_criterion} is efficient given that HSIC is estimated on the feature space.

\subsection{Memory buffer data dropout mechanism}
\label{sec:dropout}
Since the memory buffer in the mixture model continually adds incoming samples, removing several stored samples from the memory buffer is necessary in order to keep the memory size in check. Let $|\mathcal{G}|^{max}$ represent the maximum number of samples in the memory. We introduce two simple dropout mechanisms to regularize the memory capacity. The first dropout mechanism, called EEM-SW, consists of using a sliding window successively removing the initially stored samples while adding new incoming samples in the memory buffer. The second dropout mechanism, called EEM-Random, randomly drops out sets of data samples from the memory buffer ${\mathcal{G}}_j$ at $\mathcal{T}_j$.
 
\subsection{Implementation}
The whole training procedure of the evolved mixture model can be divided into three main steps~:

\noindent \textBF{Step 1 (Training phase.)} At the training step ($\mathcal{T}_j$), the current memory ${\mathcal{G}}_{j-1}$ is updated to ${\mathcal{G}}_j = {\mathcal{G}}_{j-1} \bigcup {\mathcal{D}}_{j}$. Then we train the current expert ($K$-th expert) on the memory buffer $\mathcal{G}_j$ using the loss function (Eq.~\eqref{VAEloss_eq} and Eq.~\eqref{classifierLoss_eq}).

\noindent \textBF{Step 2 (Checking the model's expansion.)} In order to avoid frequently evaluating Eq.~\eqref{expansion_criterion}, we only check the model's expansion when the current memory is full, $|{\mathcal{G}}_j| \ge |\mathcal{G}|^{max}$. To check the expansion, we calculate the HSIC measure between each expert and the data from the memory buffer, using Eq.~\eqref{hsic_measure}. Then the mixture model will add a new expert if condition~\eqref{expansion_criterion} is satisfied. For the new expert to learn statistically non-overlapping samples, we also clear up the current memory buffer ${\mathcal{G}}_j$ after the mixture's expansion.

\noindent \textBF{Step 3 (Dropout mechanism.)} This step aims to avoid memory overload. If the current memory buffer $\mathcal{G}_j$ at $\mathcal{T}_j$ is full, we drop out several samples (10 samples) from memory according to the dropout mechanism.

\renewcommand\arraystretch{1.1}
\begin{table}[tt]
    \centering
    	\caption{Classification accuracy after five indepdnent runs for various models on three datasets. * and $\dag$ denote the results cited from \cite{ContinualPrototype} and \cite{GradientTFCL}, respectively.}
\small
\setlength{\tabcolsep}{1.5mm}{\begin{tabular}{@{}l c c c @{} } 
\toprule 
Methods   &Split MNIST&Split CIFAR10 &Split CIFAR100  \\
\midrule % In-table horizontal line
	  iCARL*  &83.95 $\pm$ 0.21&37.32 $\pm$ 2.66&10.80 $\pm$ 0.37 \\
	CoPE-CE* &91.77 $\pm$ 0.87&39.73 $\pm$ 2.26&18.33 $\pm$ 1.52 \\
	CoPE*  &93.94 $\pm$ 0.20&48.92 $\pm$ 1.32&21.62 $\pm$ 0.69 \\
		ER + GMED$^\dag$ & 82.67 $\pm$ 1.90& 34.84 $\pm$ 2.20&20.93 $\pm$ 1.60   \\
	  ER$_{a}$ + GMED$^\dag$ & 82.21 $\pm$ 2.90&47.47 $\pm$ 3.20&19.60 $\pm$ 1.50  \\
	CURL*  &92.59 $\pm$ 0.66&-&- \\
	  CNDPM*  &93.23 $\pm$ 0.09&45.21 $\pm$ 0.18&20.10 $\pm$ 0.12 \\
	 \hline
	 \hline
	   EEM-SW& \textBF{96.79} $\pm$ 0.11&\textBF{58.81} $\pm$ 0.12 &\textBF{22.33} $\pm$ 0.15 \\
	   EEM-Random&96.73 $\pm$ 0.12&56.09 $\pm$ 0.15&21.78 $\pm$ 0.16  \\
\bottomrule 
\end{tabular}
\label{classification}
}
\vspace{-10pt}
\end{table}

\section{Experiments}
\label{sec;experiment}

\subsection{Experiment setting}

We consider the following TFCL benchmarks. \textBF{Split MNIST} and \textBF{Split CIFAR10} splits MNIST \cite{MNIST} and CIFAR10 \cite{CIFAR10} into five tasks, respectively, where each task consists of samples from two classes. \textBF{Split CIFAR100} divides CIFAR100 into 20 tasks and each task has 2500 examples from five classes.

\noindent \textBF{Network architecture and hyperparameters for the classifier.} We adapt ResNet-18 \cite{DeepRes} as the classifier used in Split CIFAR10 and Split CIFAR100 according to the setting from \cite{ContinualPrototype}. For Split MNIST, we adapt an MLP network, with 2 hidden layers with 400 units each \cite{ContinualPrototype}, as the classifier. We set the maximum memory size for Split MNIST, Split CIFAR10, and Split CIFAR100 as 2000, 1000 and 5000, respectively. For each training step, a model only accesses a batch of 10 samples while all previous batches are not available.

\subsection{Classification task}

In Table~\ref{classification} we evaluate EMM on Split MNIST, Split CIFAR10, and Split CIFAR100, and compare the results with several baselines including: Finetune which directly trains a classifier on the data stream, iCARL \cite{icarl}, CURL \cite{CURL}, CNDPM, CoPE \cite{ContinualPrototype}, ER + GMED and ER$_a$ + GMED \cite{GradientTFCL}, where GMED is Gradient based Memory Editing and ER is the experience replay \cite{ExperienceReplay}, while ER$_a$ is ER with data augmentation. The proposed approach outperforms not only single models such as CoPE, and GMED, but also the dynamic expansion model (CNDPM) on all three datasets.

\begin{table}[tt]
    \centering
    	\caption{Classification accuracy for 20 runs for various models on Split MImageNet.}
\small
\setlength{\tabcolsep}{5.2mm}{\begin{tabular}{@{}l cc  @{} } 
\toprule 
Methods   &Split MImageNet&Permuted
MNIST
  \\
  \midrule % In-table horizontal line
 ER$_{a}$ &25.92 $\pm$ 1.2 & 78.11 $\pm$ 0.7 \\
 ER + GMED &27.27 $\pm$ 1.8 & 78.86 $\pm$ 0.7 \\
 MIR+GMED &26.50 $\pm$ 1.3 & 79.25 $\pm$ 0.8 \\
 MIR &25.21 $\pm$ 2.2&79.13 $\pm$ 0.7 \\
	 \hline
	 \hline
	   EEM-SW&\textBF{28.90} $\pm$ 1.1&\textBF{80.32} $\pm$ 0.6  \\
	   EEM-Random&27.23 $\pm$ 1.2&80.28 $\pm$ 0.5 \\
\bottomrule 
\end{tabular}
\label{MINIImageNet_tab}
}
\vspace{-10pt}
\end{table}

We also investigate the performance of various models on the large-scale dataset, MINI-ImageNet \cite{TinyImageNet}. We split MINI-ImageNet into 20 disjoint subsets, where each subset contains samples from five classes \cite{GradientTFCL}, called  Split MImageNet. We follow the setting from \cite{GradientTFCL} where the maximum memory size is $10K$, and we implement the classifier of each expert by a slim version of ResNet-18 \cite{DeepRes}. The results provided in Table~\ref{MINIImageNet_tab}, indicate that the proposed EEM outperforms all other methods under this challenging dataset, where the other methods results are cited from \cite{GradientTFCL} and where MIR is the Maximally Interfered Retrieval.

\subsection{Ablation study}

\begin{figure}[http]
	\centering
		\includegraphics[scale=0.3]{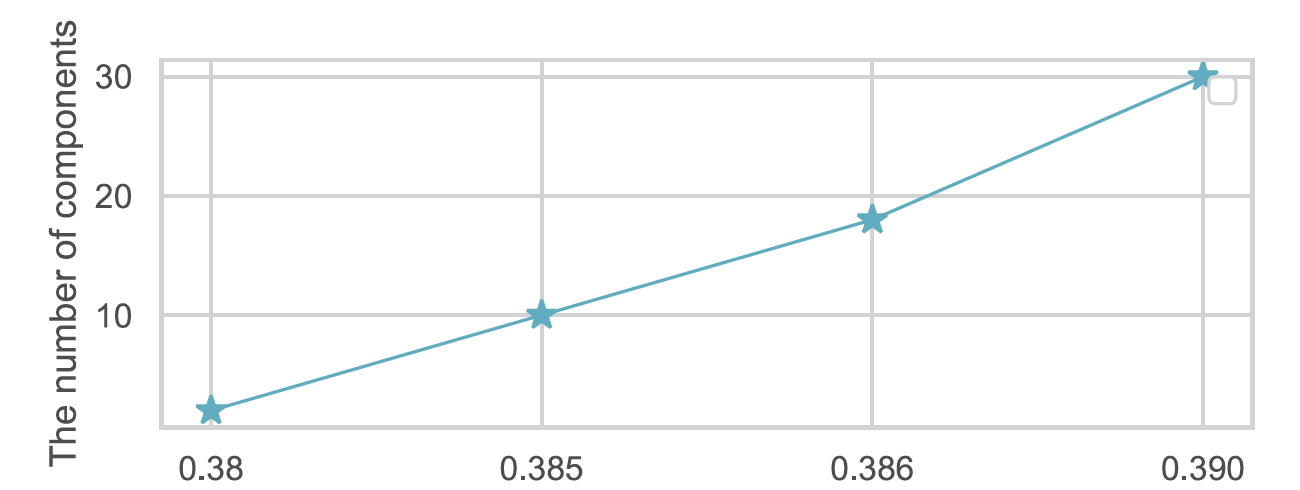} \\
		(a) No. of Components depending on the threshold $\lambda$ \\
		\includegraphics[scale=0.3]{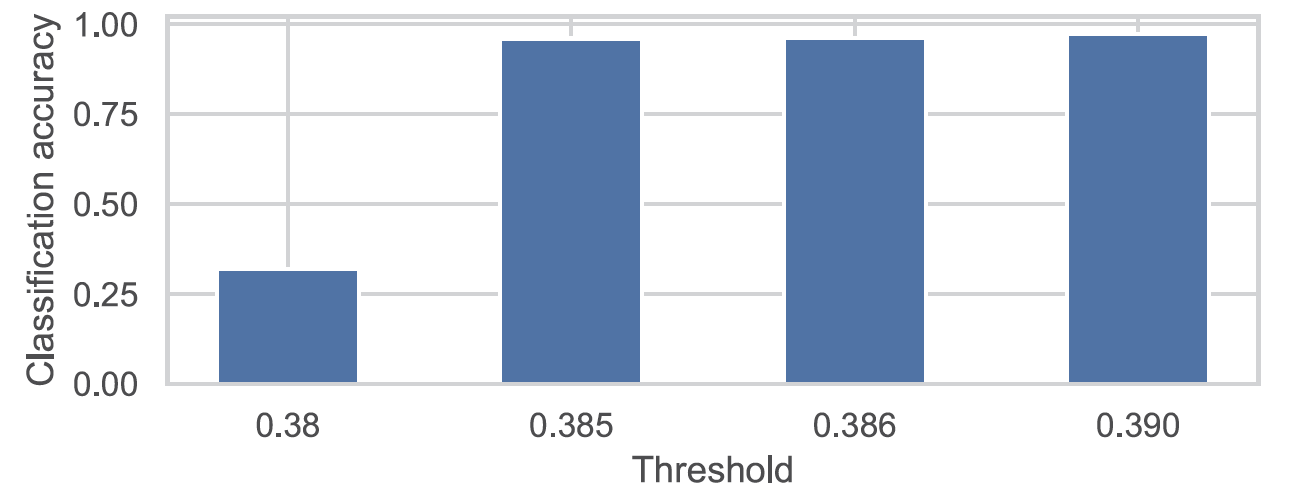}
		\\
		(b) Performance depending on the threshold $\lambda$
		\vspace{-6pt}
	\caption{Evaluation of the EMM-SW model when training on Split MNIST, 
	when varying threshold $\lambda$ in Eq.~\eqref{expansion_criterion}.}
	\label{MMD_Threshold}
	\vspace{-20pt}
\end{figure}

We perform an ablation study to investigate the performance of the proposed EEM under different hyperparameter configurations. 
The performance and the model complexity for the EEM-SW model when varying the mixture expansion threshold $\lambda$ from Eq.~\eqref{expansion_criterion} when training on Split MNIST are evaluated in Fig.~\ref{MMD_Threshold}. From Fig.~\ref{MMD_Threshold}-a, by increasing $\lambda$ leads to adding more components, but without necessary improving the classification accuracy when $\lambda$ is increased above a certain level, according to the results from Fig.~\ref{MMD_Threshold}-b.

\vspace*{-0.4cm}

\section{Conclusion}
In this paper, the Evolved Mixture Model (EMM) is proposed for learning infinite data streams without forgetting under the Task-Free Continual Learning (TFCL) setting. To address the data distribution shift in TFCL, we introduce a new mixture expansion mechanism based on the HSIC measure and also the selection of training. Finally, we perform experiments on several TFCL benchmarks, which show excellent results for the proposed approach. 

\small
\bibliographystyle{IEEEbib}
\bibliography{VAEGAN.bib}

\end{document}